\documentclass{article} 
\usepackage{iclr2025_conference,times}

\usepackage{amsmath,amsfonts,bm}

\usepackage{xcolor}
\usepackage[pdftex]{graphicx} 
\usepackage{booktabs} 
\usepackage{multirow} 
\usepackage{subcaption}
\definecolor{citeblue}{rgb}{0.21,0.49,0.74}
\usepackage[colorlinks=true,citecolor=citeblue,linkcolor=red,pagebackref,breaklinks,colorlinks]{hyperref} 

\newcommand{\red}[1]{\textcolor{red}{#1}}

\newcommand{\PAR}[1]{\vskip4pt \noindent{\bf #1~}}
\newcommand{\figtabskip}{\vskip-15pt}
\newcommand{\rebut}[1]{\textcolor{black}{#1}}

\newcommand{\shortname}{Ready-to-React}
\newcommand{\vqvae}[1]{VQ-VAE motion encoder}
\newcommand{\history}[1]{history encoder}
\newcommand{\diffusion}[1]{next latent predictor}
\newcommand{\decoder}[1]{online motion decoder}
\newcommand{\boxdataset}{DuoBox}

\newcommand{\blinterformer}{InterFormer}
\newcommand{\blcvae}{CVAE-AR}
\newcommand{\blcamdm}{CAMDM}
\newcommand{\blgpt}{T2MGPT}
\newcommand{\blduolando}{Duolando}
\newcommand{\wodec}{w/o online decoder}
\newcommand{\wodiff}{use GPT}
\newcommand{\wovq}{w/o motion encoder}
\newcommand{\usevae}{use VAE as encoder}
\newcommand{\woroot}{w/o $\rooti$ in decoder}

\newcommand{\f}{f}
\newcommand{\theagent}{\mathcal{A}}
\newcommand{\theoppo}{\mathcal{O}}
\newcommand{\agent}{\mathbf{A}}
\newcommand{\oppo}{\mathbf{O}}
\newcommand{\latent}{\mathbf{Z}}
\newcommand{\codebook}{\mathcal{C}}
\newcommand{\generator}{\mathcal{G}}
\newcommand{\agentmodel}{\mathcal{P}}
\newcommand{\past}{\mathbf{P}}
\newcommand{\rooti}{\mathbf{R}}
\newcommand{\rooto}{\mathbf{r}_\text{off}}
\newcommand{\rootd}{\mathbf{r}_\text{dir}}
\newcommand{\rootc}{\mathbf{r}_\text{dis}}
\newcommand{\posep}{\Theta_\text{pos}}
\newcommand{\poser}{\Theta_\text{rot}}
\newcommand{\posev}{\Theta_\text{vel}}
\newcommand{\oposep}{\ddot{\Theta}_\text{pos}}
\newcommand{\oposer}{\ddot{\Theta}_\text{rot}}
\newcommand{\oposev}{\ddot{\Theta}_\text{vel}}

\definecolor{agentblue}{HTML}{6E8BC5}
\definecolor{agentpink}{HTML}{D0A7C9}
\definecolor{agentgreen}{HTML}{79CC60}
\definecolor{baselinered}{HTML}{f33015}
\definecolor{oursblue}{HTML}{0600fa}
\definecolor{gtgreen}{HTML}{297e19}

\def\Tabref#1{Table~\ref{#1}}

\def\Figref#1{Figure~\ref{#1}}

\def\Secref#1{Section~\ref{#1}}

\def\eqref#1{equation~\ref{#1}}
\def\Eqref#1{Equation~\ref{#1}}

\def\1{\bm{1}}

\DeclareMathAlphabet{\mathsfit}{\encodingdefault}{\sfdefault}{m}{sl}
\SetMathAlphabet{\mathsfit}{bold}{\encodingdefault}{\sfdefault}{bx}{n}

\usepackage{hyperref}
\usepackage{url}

\usepackage[ruled]{algorithm2e} 

\title{Ready-to-React: Online Reaction Policy for Two-Character Interaction Generation}

\author{
Zhi Cen$^{1}$, Huaijin Pi$^{2}$, Sida Peng$^{1}$, Qing Shuai$^{1}$, Yujun Shen$^{3}$, Hujun Bao$^{1}$, Xiaowei Zhou$^{1}$, \\ 
\textbf{Ruizhen Hu$^{4}$\thanks{Corresponding author.}} \\
$^{1}$State Key Lab of CAD\&CG, Zhejiang University, $^{2}$The University of Hong Kong, $^{3}$Ant Group, \\
$^{4}$Shenzhen University\\
\texttt{zhicen@zju.edu.cn}, \texttt{ruizhen.hu@gmail.com}
}

\iclrfinalcopy 
\begin{document}
\maketitle

\begin{figure}[h]
    \vskip-40pt
    \begin{center}
        \includegraphics[width=\linewidth]{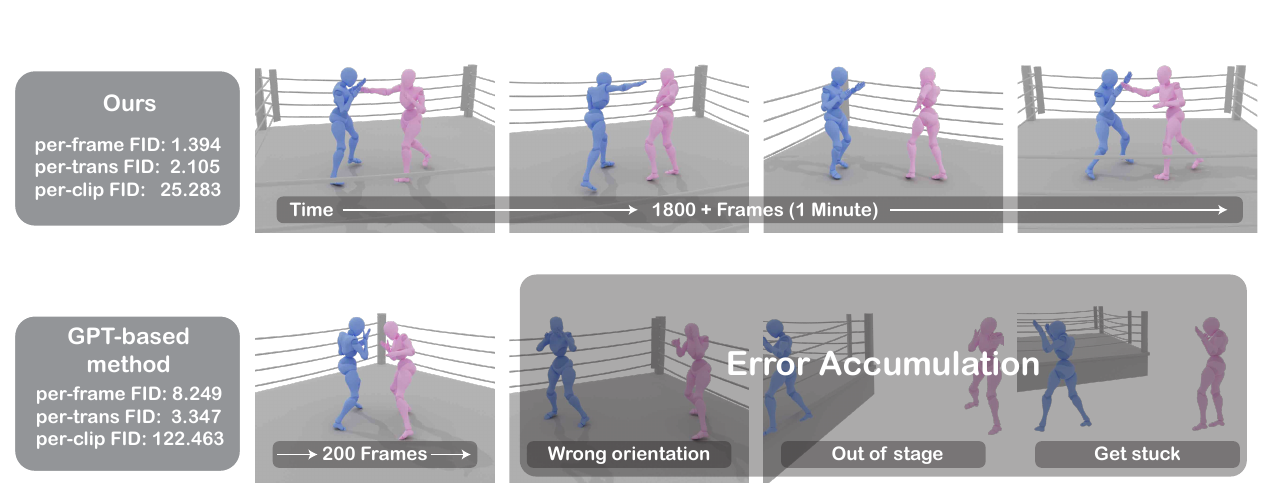}
    \end{center}
    \caption{\textbf{Demonstration of \shortname}, an \textit{online} reaction policy for two-character interaction generation on the challenging task of boxing. \shortname{} predicts the next pose of an agent by considering its own and the counterpart's historical motions. Our method can successfully generate 1800 frames of motion, whereas the GPT-based approach struggles after about 200 frames, displaying issues such as incorrect orientation, leaving the ring boundary, or freezing in place due to the accumulation of errors over time.}
    \label{fig:teaser}
    \figtabskip
\end{figure}

\begin{abstract}

This paper addresses the task of generating two-character online interactions. Previously, two main settings existed for two-character interaction generation: (1) generating one's motions based on the counterpart's complete motion sequence, and (2) jointly generating two-character motions based on specific conditions. We argue that these settings fail to model the process of real-life two-character interactions, where humans will react to their counterparts in real time and act as independent individuals. In contrast, we propose an online reaction policy, called Ready-to-React, to generate the next character pose based on past observed motions. Each character has its own reaction policy as its ``brain'', enabling them to interact like real humans in a streaming manner. Our policy is implemented by incorporating a diffusion head into an auto-regressive model, which can dynamically respond to the counterpart's motions while effectively mitigating the error accumulation throughout the generation process. We conduct comprehensive experiments using the challenging boxing task. Experimental results demonstrate that our method outperforms existing baselines and can generate extended motion sequences. Additionally, we show that our approach can be controlled by sparse signals, making it well-suited for VR and other online interactive environments. Code and data will be made publicly available at \url{https://zju3dv.github.io/ready_to_react/}.

\end{abstract}
\section{Introduction}
This paper aims to learn a reaction policy from data that can generate two-character interactions in a streaming manner.
Such a policy is essential for applications in robotics, gaming, and virtual reality, where the character needs to interact with other entities in real-time.
Generating reasonable online reactions is quite challenging, considering two key perspectives.
Firstly, the policy should dynamically adjust its own actions based on the counterpart's responses at each time step, which is essential for enabling online applications.
Secondly, downstream applications require the policy to generate natural and physically plausible motions while maintaining consistency and diversity throughout these sequences.

Previously, there were two main settings to generate two-character interaction: (1) generating one's motions based on a complete sequence of the counterpart's motion \citep{2024_duolando,2024_remos,2024_regennet}, and (2) jointly generating two-character motion sequences based on specific conditions, such as textual input \citep{2023_roleaware,2024_contactgen,2024_in2in}.
However, we argue that both settings do not model the process of real-life two-person interactions.
As humans, we dynamically produce the reaction based on the counterpart's actions at each time step, engaging in independent mind rather than sharing a collective consciousness.
To achieve two-character online interaction, it is crucial to develop a reaction policy that can be separately applied to two characters, allowing them to react to each other like real humans.

In this paper, we propose a novel reaction policy, called \shortname{}, for generating two-character interactions, as illustrated in \Figref{fig:pipeline}.
Our core innovation lies in incorporating a diffusion head into an auto-regressive model,  which can respond to the counterpart's motions in a streaming manner while ensuring the naturalness and diversity of the motions.
Specifically, we first encode the observed motions into latent vectors using a motion encoder.
Given the history motion latent of the character and the counterpart's motion, our reaction policy uses an auto-regressive model to predict a conditioning feature vector.
This vector guides a diffusion model to generate the next motion latent, which is then decoded into the next character pose.
As illustrated in \Figref{fig:teaser}, the proposed reaction policy can generate long and natural boxing sequences compared to the baseline method.

We chose boxing as the task for online two-character interaction generation considering its fast pace and frequent shifts between offense and defense. These dynamic interactions make it an ideal scenario for testing and validating our approach.
We conducted experiments to validate the effectiveness of our approach on our self-collected boxing dataset \boxdataset{}.
Our method was evaluated under three settings: (1) reactive motion generation, (2) two-character interaction generation, and (3) long-term two-character interaction generation, where it outperforms the baselines. We show that our method can generate very long motion sequences ($\sim$ 1 minute), just relying on the initial poses of the two characters.
Additionally, we carried out experiments on sparse control motion generation, demonstrating that our method is well-suited for VR online interactive settings. 
\section{Related Work}

\PAR{Single-character motion generation.}
In recent years, deep learning methods for motion synthesis have gained increasing attention \citep{2015_erd,2016_dlfcmse,2017_resrnn,2023_case,2024iclr_omnicontrol,2023iccv_hghoi,2024cvpr_textscene}.
Various techniques, including multi-layer perceptron (MLP) \citep{2017_pfnn}, mixture of experts (MoE) \citep{2018_mann}, and recurrent neural networks (RNN) \citep{2020_rmi} are employed to tackle this task. 
Additionally, to generate diverse and natural results, generative models such as conditional variational auto-encoders (cVAE) \citep{2020_motionvae}, generative adversarial networks (GAN) \citep{2022_ganimator}, and normalizing flows \citep{2020_moglow} have shown promise in addressing this challenge. 
Moreover, the success of generative pre-trained transformers (GPT) \citep{2023_t2mgpt} and diffusion models \citep{2022_mdm} further underscores their potential in this area.
Recently, \cite{2024_camdm,2024_amdm} adopt auto-regressive diffusion models to generate single-character motions.
Although our reaction policy generates one character's motion based on the other's, technically making it single-character motion generation, our work focuses on decision independence in two-character interactions.

\PAR{Reactive motion generation.}
Reactive motion generation is a subfield of human motion generation that focuses on generating human motion in response to external agents.
\cite{2022_ganreactive} introduces a semi-supervised GAN system with a part-based LSTM module to model temporal significance.
\cite{2023_interformer} employs a transformer network, enhanced by an interaction distance module using graphs. 
\cite{2024_duolando} proposes a GPT-based model to predict discrete motion tokens, enhanced by an off-policy reinforcement learning strategy. 
\cite{2024_regennet} utilizes a diffusion-based model with an explicit distance-based interaction loss. 
\cite{2024_remos} employs a diffusion-based model with a combined spatio-temporal cross-attention mechanism. 
However, \cite{2024_regennet} and \cite{2024_remos} require the complete motion sequence of the other agent, and cannot generate reactive motions online.
In contrast, our method enables online interaction generation by leveraging auto-regressive diffusion models, as inspired by \cite{2024_mar}.

\PAR{Two-character motion generation.}
While multi-character motion generation focuses on producing motion for groups with social relationships \citep{2023_socialmopred,2023_tbiformer,2023_groupdance,2023_mammos,2024_t2p},
two-character motion generation focuses on generating closer interaction between two characters.
\cite{2020_localmotionphases} proposes the local motion phase for complex, contact-rich interactions.
\cite{2021_neuralanimationlayering} combines a motion generator with task-dependent control modules. 
Both \cite{2020_localmotionphases} and \cite{2021_neuralanimationlayering} require user control signals. 
Physically-based methods \citep{2021_controlsrategies,2023_ncp,2023_maaip} ensure the physical plausibility of character interactions but still struggle to generate natural and diverse motions.
\cite{2019_ntu,2020_chi3d,2023_hi4d,2024_intergen,2024_interx} provide datasets with rich annotations that enable advancements in interaction modeling two-character scenarios.
Recent advances in text-driven two-character motion generation have introduced diffusion models \citep{2022_mdm} to enhance realism and control \citep{2023_roleaware,2024_contactgen,2024_in2in, 2024_digitallife}.
Despite these advancements, these methods jointly generate whole sequences for both characters and ignore the dynamic feedback inherent in real-life two-person interactions.

\section{Method}
\begin{figure}[t]
    \begin{center}
        \includegraphics[width=\linewidth]{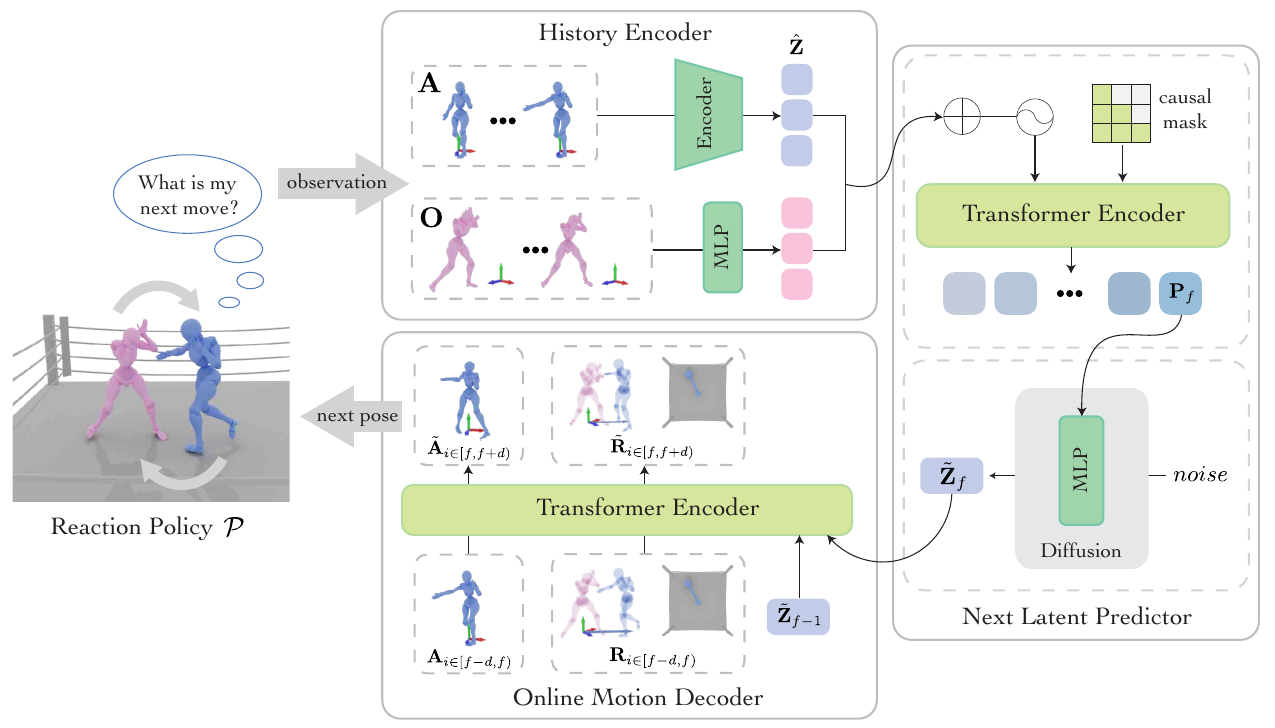}
    \end{center}
    \caption{\textbf{Pipeline overview.}
    Given a boxing scene at the leftmost figure, where the \textcolor{agentblue}{blue agent} is thinking about its next move.
    The reaction policy (\Secref{sec:policy}) follows these steps: first, based on the observations, the history encoder encodes the current state and observations; then, the next latent predictor predicts the upcoming motion latent; and finally, an online motion decoder decodes this motion latent into the actual next pose.
    The same reaction policy can be applied to the \textcolor{agentpink}{pink agent}. 
    Through a streaming process for both agents, our reaction policy enables the continuous generation of two-character motion sequences without length limit (\Secref{sec:twochar}).
    }
    \label{fig:pipeline}
\figtabskip
\end{figure}

Our aim is to develop an intelligent reaction policy capable of generating online motions that dynamically respond to a counterpart’s movements.
We begin by outlining the formulation of the one-step reaction in \Secref{sec:problem}.
Then, we introduce the design of the reaction policy in \Secref{sec:policy}. 
We explain how this policy drives the generation of two-character motions in \Secref{sec:twochar}.
In \Secref{sec:detail}, we explain the training loss and implementation details.

\subsection{Problem Formulation}\label{sec:problem}
We first introduce the problem formulation.
There are two characters, the agent $\theagent$ and the opponent $\theoppo$.
The goal is to generate the future agent's motion $\agent_{\f}$ based on the opponent's movements $\oppo_{i \in [\f-W, \f)}$ and the agent's previous motion $\agent_{i \in [\f-W, \f)}$:
\begin{equation}
    \agent_{\f} = \agentmodel(\oppo_{i \in [\f-W, \f)},~\agent_{i \in [\f-W, \f)}),
\end{equation}
where $\agentmodel$ is the generative reaction policy, and $W$ is the visible past window size.
We use the same root coordinate definition in \cite{2020_localmotionphases,2021_neuralanimationlayering}, as shown in \Figref{fig:pipeline}.
Each agent's motion $\agent_i$ is defined as:
\begin{equation}
\agent_i = \{\rooto^{\theagent} \in \mathbb{R}^2,~\rootd^{\theagent} \in \mathbb{R}^2,~\posep^{\theagent} \in \mathbb{R}^{J \times 3},~\poser^{\theagent} \in \mathbb{R}^{J \times 6},~\posev^{\theagent} \in \mathbb{R}^{J \times 3}\},
\end{equation}
where $\rooto^{\theagent}$ and $\rootd^{\theagent}$ are the horizontal trajectory (excluding the y-axis) positions and directions relative to the $(i-1)$-th frame agent's root coordinate, respectively. 
The $\posep^{\theagent}$, $\poser^{\theagent}$ and $\posev^{\theagent}$ are the positions, 6D rotations and velocities relative to the $i$-th frame agent's root coordinate, respectively.
$J$ is the number of body joints.
Each opponent's motion $\oppo_i$ is defined as:
\begin{equation}
\oppo_i = \{\oposep^{\theagent} \in \mathbb{R}^{J \times 3},~\oposer^{\theagent} \in \mathbb{R}^{J \times 6},~\oposev^{\theagent} \in \mathbb{R}^{J \times 3}\},
\end{equation}
which includes the positions, 6D rotations and velocities relative to the $i$-th frame agent's ($\theagent$'s) root coordinate, respectively.

\subsection{Reaction Policy}\label{sec:policy}
In this section, we present our reaction policy, which predicts motion in the latent space. 
As illustrated in \Figref{fig:pipeline}, the \history{} module is responsible for encoding the observations of the current state. 
Based on this historical information and the opponent's motion, the \diffusion{} module predicts the future motion latent codes. 
Finally, the \decoder{} generates the future motions using the predicted latent and the current agent's state.

\PAR{Latent motion representation.}
Converting raw data to latent space has become a popular approach in generation pipelines \citep{2022_latentdiffusion}, and VQ-VAE has been proven to be effective in learning disentangled representations of human motion data \citep{2023_t2mgpt,2024_motiongpt,2024_t2lm, 2024_sebasvq}.
In this paper, we adopt a similar architecture as in \cite{2023_t2mgpt} to encode the raw motion data into latent sequences.
Given a sequence of agent motion $\agent = \{\agent_i~|~i \in [0, f),~i \in \mathbb{Z} \}$, the VQ-VAE motion encoder encodes it to $\latent = \{\latent_i~|~i \in [0, \lfloor \frac{f}{d} \rfloor), ~i \in \mathbb{Z} \}$ with a downsampling factor $d = 4$, and each latent code $\latent_i$ is quantized through the codebook $\codebook$ to find the most similar element:
\begin{equation}
    \hat{\latent}_i=\underset{\codebook_k \in \codebook}{\text{arg min}} \left\|\latent_i-\codebook_k\right\|_2.
\end{equation}\vskip-10pt

\PAR{History encoder.}
First, we need to encode the past information of both the agent and the opponent.
To represent the agent’s past motion $\agent$, we could directly utilize the VQ-VAE motion encoder to compress them into latent variables $\hat{\latent}$.
The opponent’s historical motion $\oppo$ is also downsampled by a factor of $d = 4$ and then passed through a single-layer MLP, which encodes it into a feature vector with the dimension of $\hat{\latent}$, ensuring consistency for subsequent processing.

\PAR{Next latent predictor.}
We then predict the next motion latent in an auto-regressive manner based on the historical information. 
Our approach employs a transformer-based condition encoder to effectively capture all visible information.
This encoded data is then passed to the diffusion-based motion latent predictor, which generates the next motion latent based on the encoded conditions.

Specifically, we begin by constructing a transformer-based encoder to encode the information accessible to the reaction policy.
As illustrated in \Figref{fig:pipeline}, the transformer's input consists of the motion latent code $\hat{\latent}$ and the opponent's feature obtained from the history encoder.
As a result, each output of the transformer-based encoder $\past_f$ encapsulates the information preceding the $f$-th frame.

Next, we introduce the diffusion process, which predicts the future motion latent codes $\tilde{\latent}_{\f}$ based on $\past_\f$ at the $\f$-th frame.
We use conditional diffusion models \citep{2022_mdm, 2022_dalle2} and $\past_\f$ serves as the conditioning input, as shown in \Figref{fig:pipeline}.
We employ a single-layer MLP as the generative model $\generator$, ensuring that our model can operate in real-time.
The predicted motion latent code $\tilde{\latent}_{\f}$ is then used to generate the future motion through the \decoder{}.

Compared to previous methods that rely on predicting motion tokens' probabilities and supervising GPT models with cross-entropy loss, our approach of predicting motion latent using a diffusion-based model preserves the smooth and continuous nature of motion. This results in fewer cumulative errors and less deviation from the intended motion over time, offering greater stability and accuracy than predicting token probabilities with GPT models.

\PAR{Online motion decoder.}
A remaining problem is to decode the predicted motion latent code $\tilde{\latent}_{\f}$ into the future agent motion $\tilde{\agent}_{i \in [\f, \f +d)}$ online. 
We propose an online motion decoder that takes a few previous frames and two consecutive motion latent codes to generate the next motion frames.
As shown in \Figref{fig:pipeline}, we use a transformer as the \decoder{}.
The inputs to the \decoder{} are the past agent motions $\agent_{i \in[\f-d, \f)}$, root information $\rooti_{i \in[\f-d, \f)}$, the last motion latent code $\tilde{\latent}_{f-1}$ and current motion latent code $\tilde{\latent}_{\f}$ from the \diffusion{}.
Here, each root information $\rooti_i$ is defined as:
\begin{equation}
    \rooti_i = \{\rooto^{\theoppo} \in \mathbb{R}^2, ~\rootd^{\theoppo} \in \mathbb{R}^2,~\rootc \in \mathbb{R}^1 \},
    \label{eq:root}
\end{equation}
which includes the agent's horizontal trajectory (excluding the y-axis) positions and directions relative to the opponent's root coordinate, and the distance between the agent's root and the center of the boxing ring.
The output of the \decoder{} are the future agent motions $\tilde{\agent}_{i \in[\f, \f+d)}$ and the future root information $\tilde{\rooti}_{i \in[\f, \f+d)}$.
In contrast to VQ-VAE decoder \citep{2023_t2mgpt}, which requires the entire sequence of tokens before decoding, our method decodes motion latent into explicit motion sequences in real-time using only a few tokens and historical data, enabling online generation.

\subsection{Online Two-character Motion Generation} 
\label{sec:twochar}
Our reaction policy, as described in \Secref{sec:policy}, enables the generation of the next motion frame by leveraging both the opponent’s past motion and the agent’s own past motion. 
To implement online two-character interaction generation, we use the same reaction policy $\agentmodel$ for the two characters.

Starting with an initial input of \rebut{$s=4$} frames of poses, both characters use the policy $\agentmodel$ to generate the next $d$ frames by considering their own initial motion and the opponent’s motion. 
These generated motions are then added to a history buffer $\mathcal{H}$ with a maximum size of $W$. 
After this initial phase, both characters continuously update their predictions by incorporating the newly generated frames and the accumulated motion history. 
The interaction generation process operates in a streaming fashion. Motion that exceeds the buffer size $W$ is discarded as outdated information. 
Through this approach, both characters dynamically respond to the other, ensuring coherent and natural interactions while enabling two-character motion generation without a length limit.

\subsection{Training Loss and Implementation Details}\label{sec:detail}
The training process is divided into two stages: (1) pre-training the VQ-VAE model and (2) jointly training the \diffusion{} model and the \decoder{}.
Details about the network architecture are provided in Appendix \red{C}.

\PAR{Stage 1.}
We pre-train the VQ-VAE model following the approach in \cite{2023_t2mgpt,2019_vqvae2} for $40k$ iterations, using motion sequences cropped to 64 frames. 
The batch size is set to 128, with a codebook size = 512, codebook feature dimension = 512, and a downsampling rate of $d = 4$.
The codebook is updated with the exponential moving average (EMA) method \citep{2019_vqvae2}, as a replacement for the codebook loss.
Finally, the loss is defined as:
\begin{equation}
    \mathcal{L}_{\text{vqvae}}=\mathcal{L}_{rec} + \alpha \| sg[\hat{\latent}] - \latent \|_2^2,
\end{equation}
where $\mathcal{L}_{rec}$ is the L2 reconstruction loss, $\| sg[\hat{\latent}] - \latent \|_2^2$ is the commitment loss, the operator $sg$ refers to a stop-gradient operation, and $\alpha = 0.1$.

\PAR{Stage 2.}
Next, we train the \diffusion{} and \decoder{} jointly for $40k$ iterations while keeping the \vqvae{} fixed. 
During training, we applied causal masks (\Figref{fig:pipeline}) to ensure that the model can only access the current and previous inputs, preventing information from leaking into future time steps.
We use ground truth $\hat{\latent}_{f-1}$ instead of the predicted $\tilde{\latent}_{f-1}$ in \Figref{fig:pipeline} during training.
For this phase, motion sequences are cropped to $W = 60$ frames (2 seconds) for training. 
The batch size is set to 32, with time step $T = 1000$, and we employ DDIM \citep{2020_ddim} to sample only 50 steps during inference. 
The loss is defined as:
\begin{equation}
    \mathcal{L} = \mathcal{L}_{\text{diffusion}} + \beta \|\agent - \tilde{\agent}\|_2^2 + \gamma \|\rooti - \tilde{\rooti}\|_2^2,
\end{equation}
where $\beta = 1.0$, $\gamma = 1.0$.
$\mathcal{L}_{\text{diffusion}}$ is defined as:
\begin{equation}
    \mathcal{L}_{\text{diffusion}}=\mathrm{E}_{t \in[1, T], \mathbf{x}_0 \sim q\left(\mathbf{x}_0\right)} \left[\|\mathbf{x}_0 - \generator(\mathbf{x}_t, t, \mathbf{c})\|_2\right],
\end{equation}
where $\mathbf{x}_0 = \hat{\latent}$ is the ground truth next latent, $\generator$ is the generative model, $\mathbf{c} = \past_\f$ is the condition, and $\generator(\mathbf{x}_t, t, \mathbf{c}) = \tilde{\latent}$ is the predicted next latent.
All models are trained using the AdamW optimizer \citep{14_adam} with a learning rate of $0.0001$ on a single Nvidia RTX 4090 GPU. 
\section{Experiments} 
\subsection{Dataset, Experimental Setting, and Evaluation Metrics}
\PAR{Dataset.}
To evaluate our method, we collect a high-quality dataset, \boxdataset{}, using the OptiTrack Mocap system\footnote{\url{https://optitrack.com/}} equipped with 12 cameras. We invited three boxing enthusiasts to perform various boxing movements, recording multiple sequences of their actions. 
Please refer to Appendix \red{A} for the details of the data collection.
In total, \boxdataset{} consists of 63.4 minutes of motion data (approximately 457k frames) captured at 120 FPS.
For our experiments, we split the dataset into training (80\%) and testing (20\%) subsets, and downsample the original data to 30 FPS for training purposes. To further enrich the dataset, we apply the augmentation by swapping the roles of the agent and the opponent during training.

\PAR{Experimental setting.}
We evaluate our method in three scenarios: reactive motion generation, two-character interaction generation, and long-term two-character interaction generation.
In the reactive motion generation, we use the ground truth for the opponent's motion as input. For the two-character interaction generation, we provide only the initial 4 frames of poses for both characters. In both test scenarios, the motion of individual characters follows the procedure outlined in \Secref{sec:policy}, while the generation of two-character motions adheres to the process described in \Secref{sec:twochar}.

\PAR{Evaluation metrics.}
We evaluate the generated motion sequences using the following metrics:
(1) \textbf{Frechet Inception Distance (FID).} We follow \cite{2023_case} calculating per-frame, per-transition and per-clip FIDs. A lower FID indicates the generated motion is more similar to the real data.
(2) \textbf{Jitter.} We evaluate motion jittery following \cite{2024_gvhmr}. A closer value to the ground truth indicates better motion quality. 
(3) \textbf{Root Orient (RO).}
To assess the long-term consistency of the generated motion, we propose a new metric RO.
It calculates the percentage of frames where the facing direction between the two agents exceeds 45 degrees. This is motivated by the nature of boxing, where the athletes typically face to each other throughout the match.
Lower deviation from the ground truth means the generated motion aligns better with the expected interactive behavior.
(4) \textbf{Foot Sliding (FS).} Foot sliding \citep{2020_motionvae} measures the average sliding distance when the foot is close to the ground ($< 5 cm$). A value closer to the ground truth indicates better motion quality. Minimal foot slide may suggest that the generated motion remains stationary.
In addition, we provide the inference speed in Appendix \red{E} and motion diversity analysis in Appendix \red{F}.

\subsection{Comparison with Baselines} \label{sec:baseline}
\begin{table}[t]
\caption{
    \textbf{Comparison with baselines.} We compare our method with five baselines (\Secref{sec:baseline}) in the two scenarios: reactive motion generation and two-character motion generation.
    Among them, \textbf{bold} indicates the best results.
    $\downarrow$ means lower is better. $\rightarrow$ means closer to the real data is better.
}
\begin{center}
\resizebox{\linewidth}{!}{
    \begin{tabular}{lrrrrrrrrrrrr}
        \toprule
        \multirow{3}{*}{Methods} & \multicolumn{6}{c}{Reactive} & \multicolumn{6}{c}{Two-character} \\
        \cmidrule(r){2-7} \cmidrule(r){8-13}
        & \multicolumn{3}{c}{FID$\downarrow$} & \multirow{2}{*}{Jitter$\rightarrow$} & \multirow{2}{*}{RO$\rightarrow$} & \multirow{2}{*}{FS$\rightarrow$} & \multicolumn{3}{c}{FID$\downarrow$} & \multirow{2}{*}{Jitter$\rightarrow$} & \multirow{2}{*}{RO$\rightarrow$} & \multirow{2}{*}{FS$\rightarrow$}\\
        \cmidrule(r){2-4} \cmidrule(r){8-10}
        & Per-frame & Per-trans. & Per-clip & & & & Per-frame & Per-trans. & Per-clip \\
        \midrule
        Real            &- &- &- & 21.332 & 24.7\%     & 0.97    &- &- &- & 21.332 & 24.7\%	    & 0.97                    \\
        \midrule
        \blinterformer  & 0.724 	& 1.993 	& 15.061	    & 6.712    & 40.3\%     & 1.08    & 2.498	    & 4.590     & 47.194	    & 6.117    & 31.5\%	    & 1.05\\
        \blcvae         & 1.285 	& 4.233 	& 26.010	    & 28.519   & 46.2\%     & 1.13    & 5.405	    & 9.006   	& 92.978	    & 28.037   & 38.4\%	    & 1.09\\
        \blcamdm        & 1.606 	& 4.037	    & 28.503	    & 53.622   & 40.1\%     & 1.96    & 4.162	    & 7.488   	& 70.416	    & 53.994   & 22.6\%	    & 2.01\\
        \blgpt-online   & 1.721	    & 1.567	    & 30.159	    & 78.292   & 50.4\%	    & 2.65    & 8.249	    & 3.347 	& 122.463   	& 78.334   & 46.2\%	    & 2.63\\
        \blduolando-offline & 1.025	& 5.862	    & 13.606	    & \bf{22.544}   &\bf{30.8\%} & 3.19    & - & - & - & - & - & -\\
        \midrule
        Ours            &\bf{0.535} &\bf{0.995}	&\bf{9.599}     & 17.825   &34.7\%      &\bf{1.02} &\bf{1.394} &\bf{2.105}	&\bf{25.283}	& \bf{16.844}   &\bf{24.1\%}	&\bf{0.97}\\
        \bottomrule
    \end{tabular}
}
\end{center}
\label{tab:reactive}
\figtabskip
\end{table}

We compare our method against several baselines, including \blinterformer{} \citep{2023_interformer}, \blcvae{}, \blcamdm{} \citep{2024_camdm}, \blgpt-online{} \citep{2023_t2mgpt}, and \blduolando-offline{} \citep{2024_duolando}. Among these, \blinterformer{} is a deterministic model and lacks the ability to generate diverse results. \blcvae{} is a CVAE-based approach that we construct specifically for comparison. For \blcamdm{}, we modify its input to make it function as a reaction policy. 
\blgpt-online{} refers to decoding each newly generated token immediately into raw motion using a VQ-VAE decoder as soon as the token is produced.
\blduolando-offline \citep{2024_duolando} decodes the tokens after the entire sequence has been generated.
We retrain all these methods on the \boxdataset{} using similar training configurations for a fair comparison. Details can be found in Appendix \red{D}.

\begin{figure}[t]
    \begin{center}
        \includegraphics[width=\linewidth]{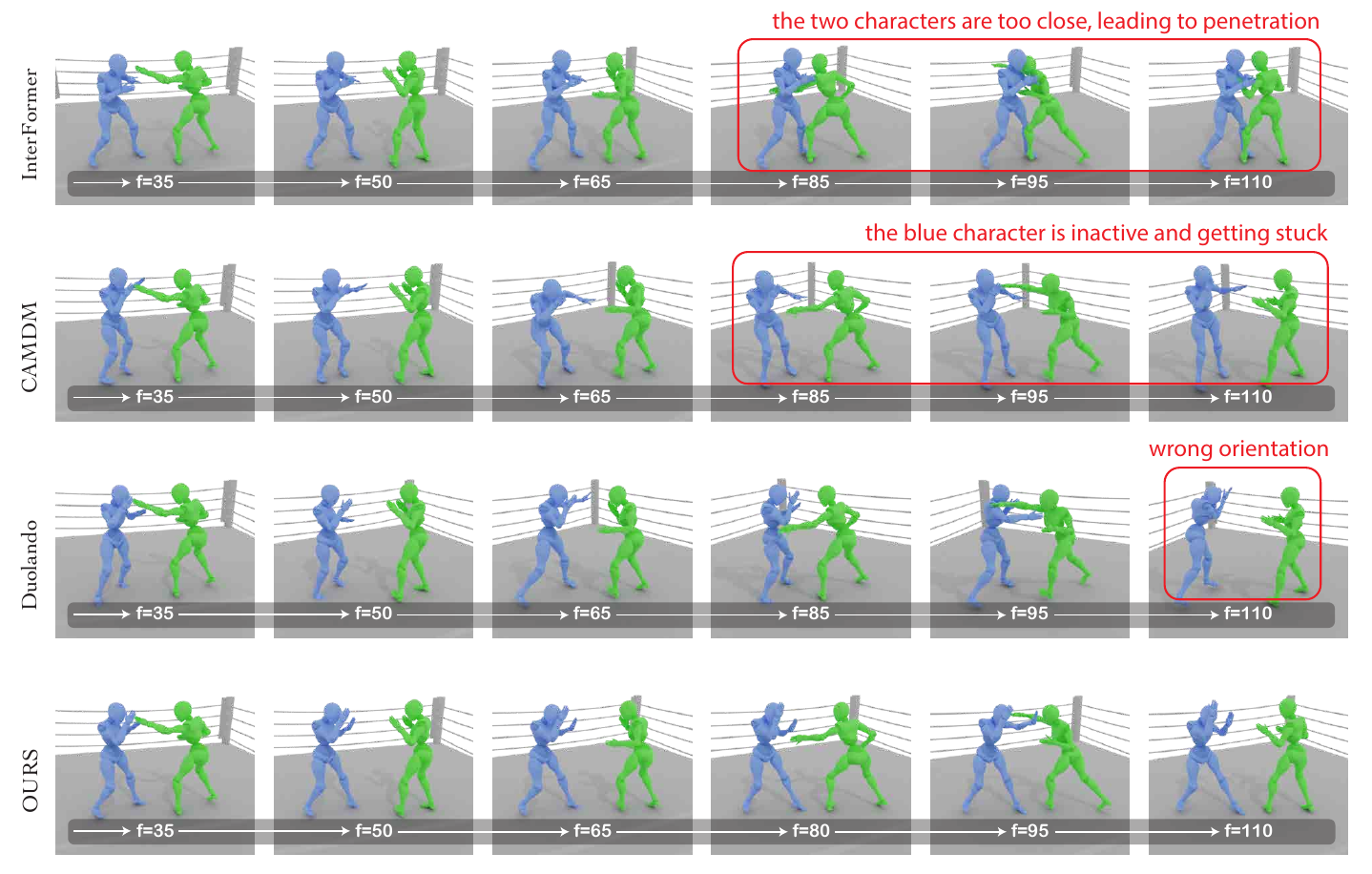}
    \end{center}
    \caption{\textbf{Qualitative results of generating reactive motions.}
    Given the same \textcolor{agentgreen}{ground truth opponent motion}, \blinterformer{} can produce reactive motion that is too close to the opponent, leading to penetration. \blcamdm{} tends to get stuck, while \blduolando{} may result in human motion with incorrect orientation after a certain period.
    }
    \label{fig:reactive}
\end{figure}

\PAR{Reactive motion generation.}
We begin by evaluating our method in the context of generating reactive motion, where the opponent's motion is provided as ground truth. The results, presented in \Tabref{tab:reactive} (left: reactive), show that our method significantly outperforms the baseline across multiple metrics, including per-frame, per-transition, and per-clip FID scores, as well as reducing foot sliding. 
Additionally, the RO of our method closely matches the performance of the offline \blduolando{} generation. Notably, the root prediction in \blduolando{} is relative to the opponent, preventing drift over time.
Among the online prediction methods, \blinterformer{} may generate motions with implausible root position and orientation.
\blgpt{}, which decodes GPT-predicted tokens online, exhibits jitter and discontinuity.
Both \blcvae{} and \blcamdm{} produce motion that is easy to get stuck over time.
We also provide qualitative results in \Figref{fig:reactive} and the supplementary materials.

\begin{figure}[t]
    \begin{center}
        \includegraphics[width=\linewidth]{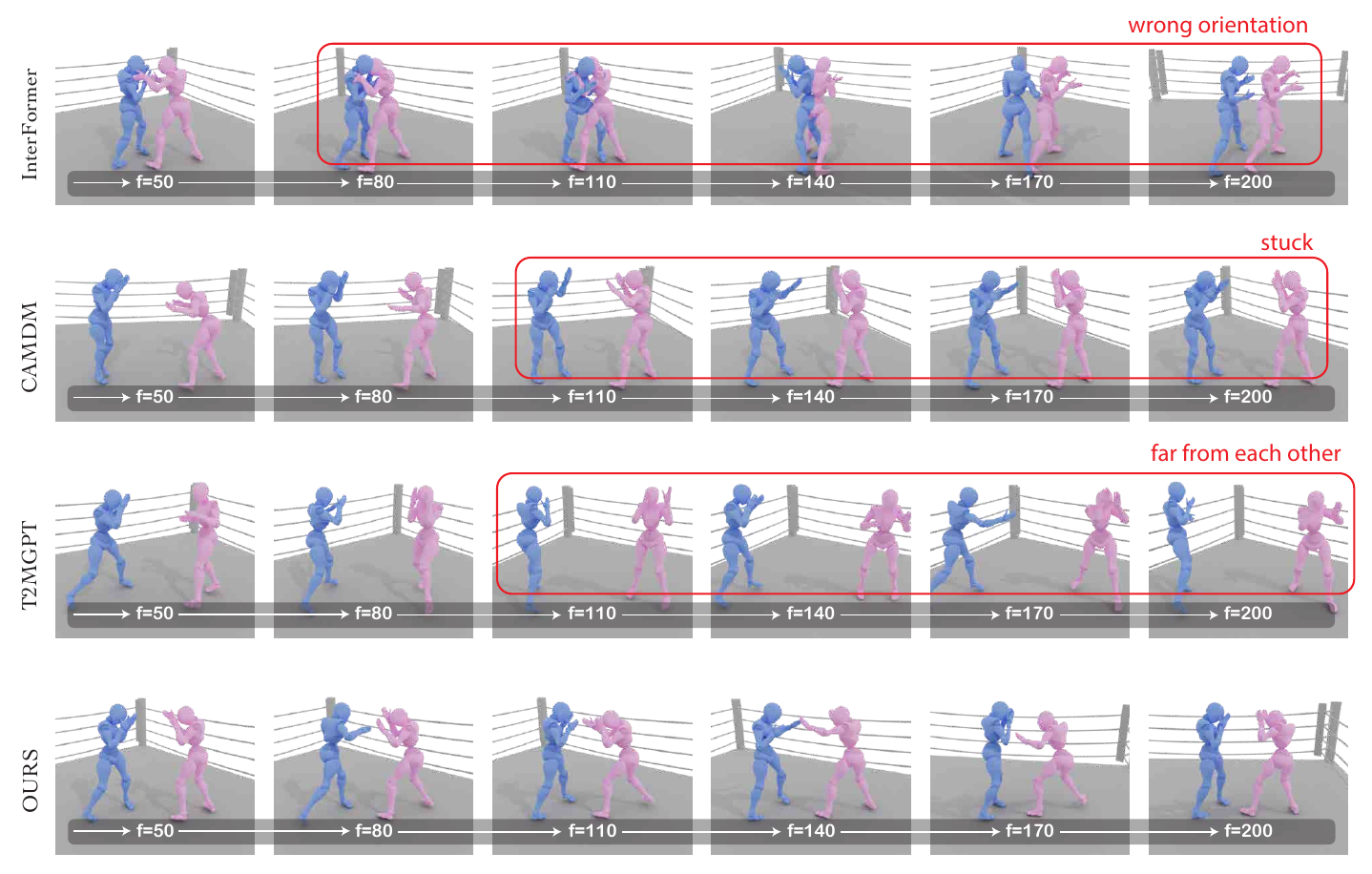}
    \end{center}
    \caption{\textbf{Qualitative results of generating two-character motions.}
    Given the same initial four frames for both characters, \blinterformer{} tends to produce human motion with incorrect orientation. \blcamdm{} often results in the characters getting stuck, while \blgpt{} can cause the two characters to drift apart due to accumulated errors.
    }
    \label{fig:twoagent}
\end{figure}

\PAR{Two-character interaction generation.}
Our method enables the simultaneous generation of motion for both agents. Starting with the first four frames, each agent's subsequent motion is generated by leveraging the interaction between their own and their opponent's past motions. 
In contrast, \blduolando-offline{} cannot generate both agents' motions simultaneously, as it predicts the root position relative to the opponent's at the same frame and requires future information through a looking-ahead mechanism. Moreover, it relies on having all tokens available before decoding, preventing online generation. 
As shown in \Tabref{tab:reactive} (right: two-character), our method significantly outperforms the baseline across all metrics in this more complex scenario.
We also provide qualitative results in \Figref{fig:twoagent} and the supplementary materials.

\begin{table}[t]
\caption{\textbf{Quantitative results of long-term two-character motion generation.} We compare our method with four baselines (\Secref{sec:baseline}).
    The generated motion lengths are set to 1800 frames.}
\begin{center}
\resizebox{0.7\linewidth}{!}{
    \begin{tabular}{lrrrrrr}
        \toprule
        \multirow{2}{*}{Methods} & \multicolumn{3}{c}{FID $\downarrow$} & \multirow{2}{*}{Jitter $\rightarrow$} & \multirow{2}{*}{RO $\rightarrow$} & \multirow{2}{*}{FS $\rightarrow$} \\
        \cmidrule(r){2-4}
        & Per-frame & Per-transition & Per-clip \\
        \midrule
        Real &- &- &- &21.332 & 24.7\%	    & 0.97  \\
        \midrule        
        \blinterformer  & 5.628	&	5.936	&	87.697	&    4.784   &	44.6\%	&	\bf{0.90} \\
        \blcvae         & 7.325	&	11.717	&	110.458	&   18.107   &	36.7\%  &	0.75 \\
        \blcamdm        & 7.557	&	13.465	&	109.654	&   25.237   &	13.7\%	&	0.86 \\
        \blgpt-online   & 21.27	&	4.457	&	273.086	&   72.481   &	70.4\%	&	2.34 \\
        \midrule
        Ours            & \bf{2.388}	&	\bf{2.375}	&	\bf{36.755}	&   \bf{19.204}   &	\bf{31.3\%}	&	0.83 \\
        \bottomrule
    \end{tabular}
}
\end{center}
\label{tab:long}
\figtabskip
\end{table}
\begin{figure}[t]
    \begin{center}
        \includegraphics[width=\linewidth]{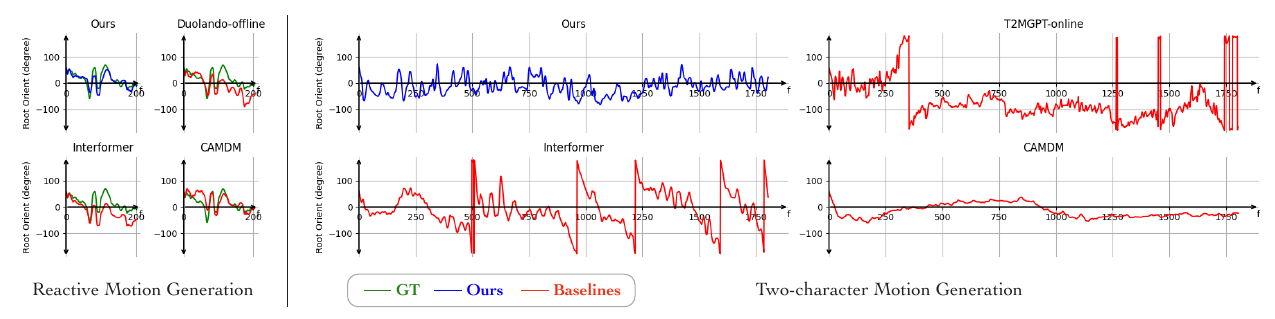}
    \end{center}
    \caption{\textbf{Visualization of the face direction relative to time.}
    We compare our method with baselines in two scenarios described in \Secref{sec:baseline}. The x-axis represents the frame number \f, while the y-axis shows the angle between the facing directions of the two characters (in degrees). An angle of $0^{\circ}$ indicates that the two agents are facing each other, whereas $\pm 180^{\circ}$ means they are facing away from each other. The \textcolor{gtgreen}{green lines} represent the ground truth, the \textcolor{oursblue}{blue lines} represent our method, and the \textcolor{baselinered}{red lines} represent the baselines.
    }
    \label{fig:rootorient}
    \figtabskip
\end{figure}

\PAR{Long-term two-character interaction generation.}
We demonstrate our method's ability to generate extended sequences of two-character motion, highlighting its reduced error accumulation and superior motion quality over long durations.
Our method is capable of generating two-character motion for 1800 frames.
The results are shown in \Tabref{tab:long}. The FID of ours is much better than all the baselines.
To show the effect of error accumulation, we plot the face direction relative to time in \Figref{fig:rootorient}. \blgpt-online and \blinterformer{} will quickly generate motions that face away from each other after some time, and \blcamdm{} generates motions that are over-smoothed.
Please refer to the supplementary material for more visualizations.

\subsection{Ablation Study} \label{sec:ablation}
\begin{table}[t]
\caption{
    \textbf{Ablation study.} We compare our method with five variants to validate our main design choices (please refer to \Secref{sec:ablation} for details).
    Among them, \textbf{bold} indicates the best results.
    $\downarrow$ means lower is better. $\rightarrow$ means closer to the real data is better.
    }
\begin{center}
\resizebox{\linewidth}{!}{
    \begin{tabular}{lrrrrrrrrrr}
        \toprule
        \multirow{3}{*}{Methods} & \multicolumn{5}{c}{Reactive} & \multicolumn{5}{c}{Two-character} \\
        \cmidrule(r){2-6} \cmidrule(r){7-11}
        & \multicolumn{3}{c}{FID$\downarrow$} & \multirow{2}{*}{RO$\rightarrow$} & \multirow{2}{*}{FS$\rightarrow$} & \multicolumn{3}{c}{FID$\downarrow$} & \multirow{2}{*}{RO$\rightarrow$} & \multirow{2}{*}{FS$\rightarrow$}\\
        \cmidrule(r){2-4} \cmidrule(r){7-9}
        & Per-frame & Per-trans. & Per-clip & & & Per-frame & Per-trans. & Per-clip \\
        \midrule
        Real        &-     &-     &-   & 24.7\%     & 0.97      &- &- &-    & 24.7\%     & 0.97 \\
        \midrule
        \usevae     &   0.525	&	1.290	&	10.423		    &	36.5\%	&	1.00	&	1.446	&	3.002	&	28.268	&	22.2\%	&	1.01    \\
        \wovq       &   0.693	&	1.339	&	24.860		    &	54.0\%	&	1.48	&	7.203   &	3.012	&	126.991	&	52.3\%	&	1.47    \\
        \wodiff     &   0.892   &	1.781	&	16.212		    &	37.4\%	&\bf{0.99}	&	2.418	&	3.536	&	41.144	&	20.9\%	&	\bf{0.97}    \\
        \wodec      &   5.215	&	11.020	&	78.933		    &	44.9\%	&	1.06	&	11.279	&	23.189	&	157.253	&	16.0\%	&   0.91    \\
        \woroot     &   \bf{0.496}	&	1.074	&	10.315		&	43.3\%	&	0.93	&	3.370	&	2.275	&	54.503	&	38.2\%	&	0.92\\
        \midrule
        Ours        &   0.535     &\bf{0.995}     &\bf{9.5998}  &\bf{34.7\%}	&1.02	    &\bf{1.394}	&\bf{2.105}	&\bf{25.283}  &\bf{24.1\%}	&\bf{0.97}  \\
        \bottomrule
    \end{tabular}
}
\end{center}
\label{tab:ablation}
\figtabskip
\end{table}
As shown in \Tabref{tab:ablation}, we compare our method with five main ablated versions:
(1) \textbf{\usevae.}
To demonstrate the stability of our method with different motion latent encodings, we replace the VQ-VAE with a VAE as the motion encoder. As shown in the table, using VAE as the motion encoder does not significantly affect the results.
(2) \textbf{\wovq.}
To highlight the importance of predicting motion latent codes rather than raw motions, we remove the \vqvae{} and directly predicted the pose sequence. The results show that directly predicting the raw pose sequence significantly degrades the motion quality.
(3) \textbf{\wodiff.}
We replace the diffusion model with GPT to predict the next token probabilities. The results show a decline in motion quality, with noticeably worse FID scores.
(4) \textbf{\wodec.}
To validate the necessity of training a new online motion decoder, we directly apply the VQ-VAE decoder to decode the motion latent codes into motion sequences. 
Using the VQ-VAE decoder at each step results in discontinuous motion, which in turn leads to a worse FID score.
(5) \textbf{\woroot.}
We remove the root sequence $\rooti$ in \Eqref{eq:root} as the input to the \decoder{}. Without $\rooti$, the model can easily predict motions with the wrong root facing direction.

In summary, we demonstrate the importance of different components in our model. More ablation studies and visual results can be found in Appendix \red{G} and the supplementary materials.
\section{Application: Generating Reactive Motion with Sparse Signals}
Introducing sparse control is essential for making our method practical in real-world applications, particularly in VR online interactive environments. In these settings, capturing detailed and dense input data can be challenging due to hardware limitations, computational costs, or user comfort. Sparse control addresses this by allowing the system to generate high-quality motion based on minimal input signals.

To demonstrate that our method is well-suited for VR online interactive environments, we also conducted experiments showing that it can be controlled by sparse signals.
The sparse signals are the head and two-hand positions and rotations relative to the previous frame's agent root coordinate.
To enable the controlling feature, we retrain the reaction policy by adding the sparse signals as conditions to the two transformer models in \Figref{fig:pipeline}. The loss and other training settings remain unchanged.
We evaluate the quality of the generated motion using FID scores, motion jitter, foot sliding, and position and rotation errors to highlight the controllability of our approach.
We compare our method with \blcamdm~\citep{2024_camdm}, an auto-regressive method that generates diverse motions based on control signals. The results, presented in \Tabref{tab:control}, show that our method consistently outperforms the baseline across all evaluated metrics.
We also provide qualitative results in \Figref{fig:sparse} and in the supplementary materials.

\begin{table}[t]
\caption{
    \textbf{Quantitative results of generating reactive motion from sparse signals.}
    We compare our method with \blcamdm.
    Among them, \textbf{bold} indicates the best results.
    $\downarrow$ means lower is better. $\rightarrow$ means closer to the real data is better.
    Our method outperforms the baseline in terms of all metrics.
    }
\begin{center}
\resizebox{0.8\linewidth}{!}{
    \begin{tabular}{lrrrrrrr}
        \toprule
        \multirow{2}{*}{Methods} & \multicolumn{3}{c}{FID $\downarrow$} & \multirow{2}{*}{Jitter $\rightarrow$} & \multirow{2}{*}{FS $\rightarrow$} & \multirow{2}{*}{Pos. Err. $\downarrow$} & \multirow{2}{*}{Rot. Err. $\downarrow$} \\
        \cmidrule(r){2-4}
        & Per-frame & Per-transition & Per-clip \\
        \midrule
        Real &- &- &- &21.332 & 0.97 & - & -  \\
        \midrule
        \blcamdm & 0.697 & 1.506 & 15.169 & 47.229 & 2.25 & 14.52 & 22.40 \\
        Ours & \bf{0.249} & \bf{0.263} & \bf{4.086} & \bf{21.163} & \bf{1.06} & \bf{2.72} & \bf{4.39} \\
        \bottomrule
    \end{tabular}
}
\end{center}
\label{tab:control}
\figtabskip
\end{table}
\begin{figure}[t]
    \begin{center}
        \includegraphics[width=\linewidth]{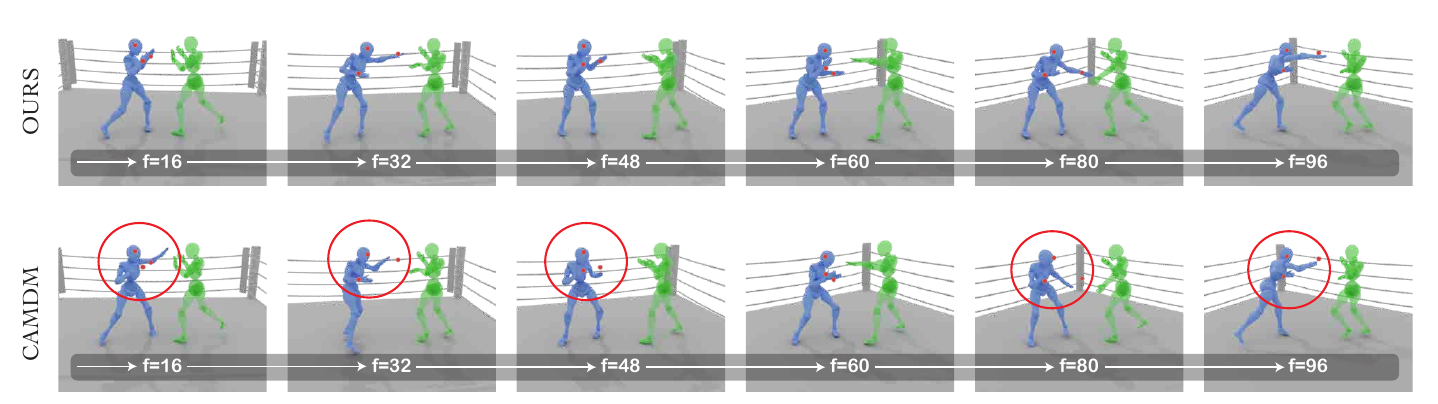}
    \end{center}
    \caption{
        \textbf{Qualitative results of generating reactive motions from sparse signals.} We compare our method with \blcamdm. Our approach successfully generates realistic motion while effectively adhering to the sparse signals (annotated by \red{red dots} in the figures). In contrast, \blcamdm{} struggles to achieve the same level of responsiveness and accuracy, as shown in the \red{red circles}.
    }
    \label{fig:sparse}
    \figtabskip
\end{figure}
\section{Discussion and Conclusion}

In this work, we have introduced a novel reaction policy that can be applied to two-character boxing interaction generation.
The reaction policy includes a diffusion-based predictor for forecasting the next motion latent, paired with an online motion decoder optimized for the online generation. Our method effectively reduces error accumulation, enables real-time online generation, and can produce extended motion sequences. Additionally, it offers controllability through sparse signals, making it well-suited for online interactive environments.

However, several limitations to our approach should be addressed in future work. First, the current method is primarily designed for interactions between two individuals, and extending it to handle multi-person scenarios remains a challenge. Second, our method does not account for interactions with the environment or objects, which are essential for many real-world applications. Overcoming these limitations will be crucial for making the method more versatile and applicable to a wider range of scenarios.

\section*{Acknowledgements}
This work was partially supported by the NSFC (No.~62322207, No.~62172364, No.~62402427), Ant Group Research Fund and Information Technology Center and State Key Lab of CAD\&CG, Zhejiang University. We also acknowledge the EasyVolcap \citep{easyvolcap} codebase.

\clearpage

\clearpage
\bibliography{iclr2025_conference}
\bibliographystyle{iclr2025_conference}

\end{document}